\documentclass[sigconf,nonacm]{acmart}
\pdfoutput=1

\usepackage[T1]{fontenc}   
\usepackage{booktabs}      

\usepackage{bbm}
\usepackage{amsmath}
\usepackage[skip=1pt]{subcaption}
\newcommand{\ind}[1]{\mathbbm{1}_{#1}}

\title{Transformers à Grande Vitesse}
\subtitle{Massively parallel real-time predictions of train delay propagation}

\author{Farid Arthaud}
\email{farto@csail.mit.edu}
\affiliation{
	\institution{Ecole Normale Supérieure}
	\streetaddress{45 rue d'Ulm}
	\city{Paris}
	\country{France}
	\postcode{75005}
}
\affiliation{
	\institution{SNCF Réseau}
	\streetaddress{10-12 rue Camille Moke}
	\city{Saint-Denis}
	\country{France}
	\postcode{93200}
}

\author{Guillaume Lecoeur}
\email{ext.guillaume.lecoeur@reseau.sncf.fr}
\author{Alban Pierre}
\email{ext.alban.pierre@reseau.sncf.fr}
\affiliation{
	\institution{SNCF Réseau}
	\streetaddress{10-12 rue Camille Moke}
	\city{Saint-Denis}
	\country{France}
	\postcode{93200}
}


\begin{document}
\begin{abstract}
Robust travel time predictions are of prime importance in managing any
transportation infrastructure, and particularly in rail networks where they
have major impacts both on traffic regulation and passenger satisfaction.
We aim at predicting the travel time of trains on rail sections at the scale of
an entire rail network in real-time, by estimating trains' delays relative
to a theoretical circulation plan.

Predicting the evolution of a given train's delay is a uniquely hard problem,
distinct from mainstream road traffic forecasting problems, since it involves
several hard-to-model phenomena: train spacing, station congestion and
heterogeneous rolling stock among others.
We first offer empirical evidence of the previously unexplored phenomenon of
\emph{delay propagation} at the scale of a railway network, leading to delays
being amplified by interactions between trains and the network's physical
limitations.

We then contribute a novel technique using the transformer architecture and
pre-trained embeddings to make real-time massively parallel predictions for
train delays at the scale of the whole rail network (over 3000 trains at peak
hours, making predictions at an average horizon of 70 minutes).
Our approach yields very positive results on real-world data when compared to
currently-used and experimental prediction techniques.
\end{abstract}

\begin{teaserfigure}
  \centering
  \subcaptionbox{8:30 AM}{\includegraphics[width=.21\textwidth]{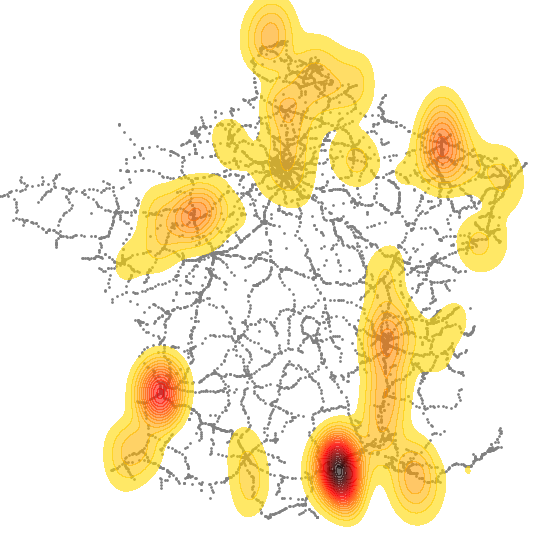}} \hspace{.025\textwidth}
  \subcaptionbox{12:30 AM}{\includegraphics[width=.21\textwidth]{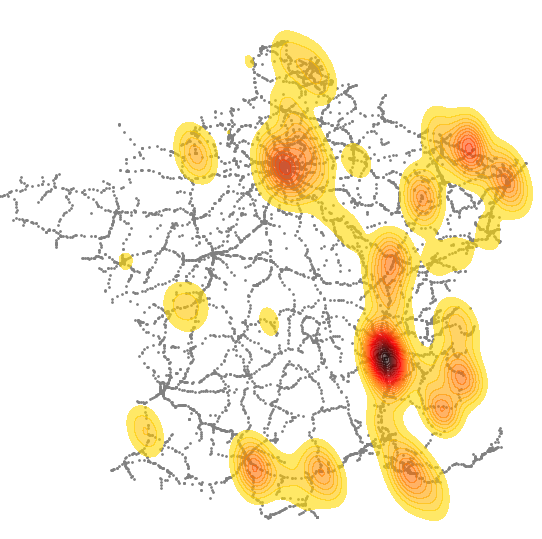}} \hspace{.025\textwidth}
  \subcaptionbox{3:00 PM}{\includegraphics[width=.21\textwidth]{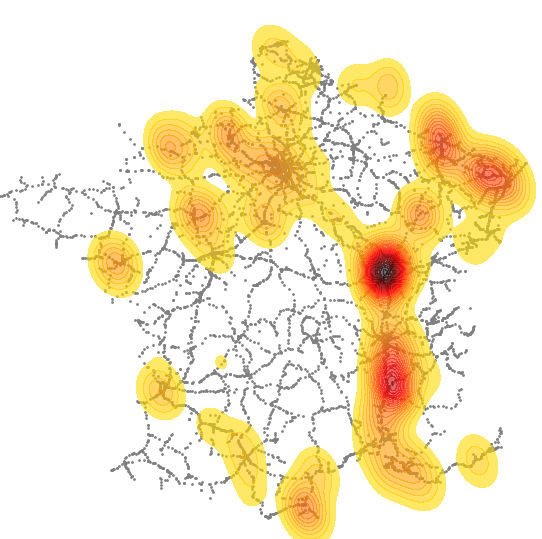}} \hspace{.025\textwidth}
  \subcaptionbox{7:30 PM}{\includegraphics[width=.21\textwidth]{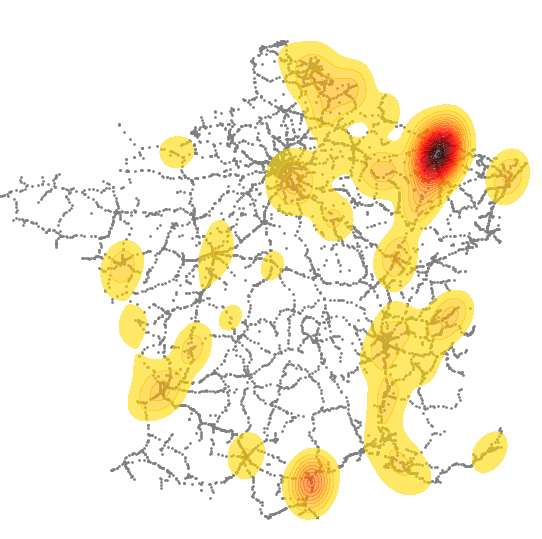}}
  \caption{Gaussian density model of train delays on the NRN on March
  5\textsuperscript{th} 2019, demonstrating delay propagation}
  \Description{Four maps of the density of delays on the national rail network
  as colored outlines, showing a dark spot moving from Montpellier to Metz.}
  \label{fig:propa}
\end{teaserfigure}

\maketitle

\section{Introduction}
Delay prediction is one of the most fundamental and oldest computational problems
in railway operations.
They are important both to infrastructure operators, as predicting delays can aid
in real-time traffic regulation, and to passenger rail operators to improve
passenger information and therefore
satisfaction~\citep{tiong2023review,yaghini2013railway}.

Moreover, the growing scale and interconnectedness of rail networks, as well as
the wealth of delay data accumulated by rail companies can make it hard to
isolate specific phenomenon which cause delays.
Besides \emph{extrinsic delays}, the phenomena causing \emph{delay propagation}
have been of particular interest in the literature in recent
years~\citep{goverde2010delay,bosscha2016big,harrod2019closed}.
Predicting the evolution of a given train's delay is a uniquely hard problem,
distinct from mainstream road traffic forecasting problems, since it involves
several hard-to-model phenomena: train spacing, station congestion and
heterogeneous rolling stock among others, which can all lead to delay
propagation.

We give a succinct description of the \textbf{delay propagation} phenomenon in
Section~\ref{sec:prop}, wherein late trains will cascade additional delays for
other trains.
Existing literature often treats these phenomena at a local scale, or isolates
each type of phenomenon causing delay propagation (which we outline in
Section~\ref{sec:prop}).
As explained in Section~\ref{sec:prop} and Section~\ref{sec:tran}, we treat
all causes of delay propagation together, and use massive network-scale
predictions to encompass all aspects of delay propagation.

Real-time delay predictions for medium-term horizons are useful for a variety of
reasons for railway operators: they feed passenger information systems and they
allow network regulators to make more informed decisions.
Moreover, the study of delay propagation can also help network operators
understand where extraneous delays are created, and help fluidify the network.
These real-world requirements translate into a prediction horizon of the order of
magnitude of several dozens of minutes, a scale precise enough to make
predictions in each train station and major junction visited by a train, and a
prediction computation time below a minute (for real-time use): our approach
satisfies all of these real-world usage criteria.

To this end we propose a novel approach based on the transformer architecture and
pre-trained embeddings, described in Section~\ref{sec:tran}, providing real-time
predictions for all trains on the French national rail network by taking into
account propagation at the time scale of \( 70 \) minutes on average.
The main idea of our work is to make predictions in parallel at the network scale,
taking into account all trains' states when making predictions for a single
train, along with an extensive history of every train's itinerary.
We call our method `massively parallel' for this reason: in contrast with
existing work, all propagation paths between all trains are considered by our
architecture.
We finally present our results and compare them experimentally both to currently
deployed techniques (our baseline) and a statistical Bayesian network technique
in Section~\ref{sec:res}.

\section{Related work}\label{sec:related}
Most work using neural networks for traffic forecasting in transportation
networks so far has been focused on road traffic.
These tasks typically model average speed on highways (such as the METR-LA and
PEMS-BAY datasets) or traffic flow along city streets (BJF and
BRF,~\citet{zhang2020spatio}); however these variables are often ill-defined in a
railway context: 96.7\% of edges of our network never contain more than 4
trains simultaneously, and during peak hour over 97.3\% of edges contain no
trains.
Railways also introduce many constraints that don't exist for road traffic:
incidents have much more propensity to completely halt flow of traffic on most
one-way tracks, train spacing strongly limits traffic density, and platform
occupancy constraints in stations create unavoidable congestion at specific
nodes.
The sparsity of train traffic along with the branching structure of railway
networks thus make it impossible to apply approaches designed for road traffic,
which have a macroscopic view of traffic in contrast with the vehicle-by-vehicle
basis required for rail traffic.
A closer task to ours is estimating single vehicle travel time within a road
network based on congestion information in a city's
network~\citep{hong2020heteta, tran2020deeptrans}.
However, the data used for inference is very different from our rail setting:
congestion generally is not the determining factor for delays in rail traffic,
and as outlined above is often not well-defined in our problem.

Machine learning has been used to improve traffic predictions for trains.
A variety of statistical methods have been suggested to model delay propagation,
for example using timed event graphs~\citep{goverde2010delay}, fuzzy Petri
networks~\citep{milinkovic2013fuzzy}, activity
graphs~\citep{buker2012stochastic}, random forests~\citep{li2021near}, Bayesian
networks~\citep{chaker2020modeliser}, and integer
programming~\citep{dollevoet2018delay}.
\citet{spanninger2022review} and~\citet{tiong2023review} offer
detailed and complete reviews of the area, which is not in the scope of this
publication.
These models are however constrained by scale, modeling propagation from an
individual train to another, most often on a single railway line.
At the scale of an entire network, as is our case, massive delay propagation as
demonstrated in Section~\ref{sec:prop} would entail a lot of uncertainty using
train-to-train propagation as errors would accumulate.
Most work so far referring to `delay propagation' limits itself to propagation
between the different stops of a single train, or along a single railway line
between trains following each other; whereas we account for interactions between
all trains, introducing many more propagation paths.
Moreover, these works choose to model each propagation phenomena separately,
making way for omissions or imprecisions, whereas our model summarizes all
forms of propagation in a single prediction scheme.

Feed-forward neural networks have previously been used to forecast delays for
individual trains~\citep{yaghini2013railway, markovic2015analyzing,
chapuis2017arrival, oneto2018train, huang2020modeling, hauck2020data}, using
variables such as origin and destination of the train and past observed delays.
These models are often relatively simple (less than \( 10 \)k parameters), and
don't account for delay propagation between trains.
\citet{bosscha2016big} uses recurrent neural networks for single-train delay
predictions, incorporating a form of delay propagation between trains by
tracking how many trains departed at the same time as a given train.
This single variable however fails to capture large-scale delay propagation
phenomena we are interested in.
\citet{li2022prediction} study propagation at multi-line stations using
convolutional neural networks, and contains a partial review of the area of delay
forecasting.
Their approach focuses on delay propagation at a single railway station: as
detailed in Section~\ref{sec:prop}, our approach encompasses many other modes of
propagation that may occur outside of train stations, for example between trains
which do not share a station.
One of the main benefits of our approach is the self-attention mechanism of the
transformer architecture, which is not present in feed-forward networks and which
we use to capture usual propagation paths between trains.
As our task is very different from these works in scale but also in the
propagation mechanisms we aim to predict (see Section~\ref{sec:prop}),
experimental comparisons would be ill-defined: scaling up existing approaches or
scaling down our approach to single trains, stations or lines would require
significant modifications to both.

Our work uses the transformer architecture~\citep{vaswani2017attention} to make
delay predictions, a neural network architecture based on a self-attention
mechanism first proposed for machine translation.
It has since been declined to other neural language processing (NLP) tasks such
as language modeling~\citep{devlin2019bert}; but also different domains such as
computer vision tasks~\citep{han2021survey} and time series
forecasting~\citep{li2019enhancing}.
Recent work on road traffic forecasting tasks has also used the
transformer architecture~\citep{cai2020traffic, xu2020spatialtemporal}, using
attention for spatial and temporal domains.
As above, these are not applicable to the rail domain, but validate the
relevance of self-attention for propagation forecasting.
\citet{cai2020traffic} use transformer inputs for traffic at different times,
and deal with the spatial dimension using a graph neural network layer: this is
the dual to our approach, where time is encoded as components of the tokens fed
to the transformer encoder, and spatiality is captured through the different
tokens (each corresponding to a train, see details in Section~\ref{sec:tran}).
\citet{xu2020spatialtemporal} alternate spatial and temporal encoders, also using
graph neural networks to encode spatial data.
Once more, our approach avoids the use of graph neural networks (as well as
positional encodings in the transformer) by leveraging the fact that spatial
features can be passed along discrete train positions, which is not the case in
road traffic where individual vehicles are too numerous and hard to locate for
such an approach.
To the best of our knowledge, no prior work uses the transformer architecture for
delay propagation forecasting at the scale of an entire rail network.

\section{Delay propagation}\label{sec:prop}
Making predictions for train delays relative to a theoretical circulation plan,
given the state of the entire railway network requires to account for a variety
of extrinsic and intrinsic factors described below.
We distinguish \emph{primary delays}, introduced organically for on-schedule
trains, from \emph{secondary delays} (sometimes called \emph{knock-on
delays}) which are propagated from or amplified by existing primary delays or
degraded circulation conditions.

The following is a coarse and non-exhaustive categorization of the main sources
of delays observed on a typical railway network,
\begin{enumerate}
\item \textbf{Extrinsic incidents}: incidents unrelated to previous incidents or
	the theoretical circulation plan are a primary and unpredictable source
	of lateness, with 1,250 occurring on average each day on the French
	national railway network.
	Examples are animals being present on the tracks or a train breaking
	down.
\item \textbf{Inconsistent circulation plan}: imperfections in the circulation
plan are also a primary source of delays. This is most often due to an
optimistic modelisation of real-world conditions, such as acceleration curves,
maximum speeds or passenger loading time. This is visible in our data through
recurring biases in delays at specific edges in the network.
\item \textbf{Race conditions}: a train can accumulate delay based on race
conditions at certain railway nodes such as junctions or stations.
This is a secondary source of lateness, since it requires for another train to be
late in order to occur.
Examples include waiting at a signal when it is reached later than scheduled,
waiting before a station when a train's allotted platform has become occupied,
and traveling at a reduced speed when running behind a slower rolling stock.
\item\label{pnt:regu} \textbf{Traffic regulation decisions}: traffic regulation
is done entirely by humans and based on experience foremost.
This means that agents' discretion dictates how recovery from an incident is
carried out, and human error or misperception might introduce flaws in
regulation.
This leads to sub-optimal traffic conditions, or exacerbated secondary delays
which are hard to predict due to the added human element.
\item \textbf{Traffic density}: at peak hours, major arteries on the railway
	network
are saturated, meaning train spacing is at a minimum all along the line. If a
train comes too close to the train in front of it, automatic signals (KVB in
France) will slow it down,
leading to accordion effects all along the line.
This fragile equilibrium is easily disturbed especially at the end of
high-speed sections or at junctions along these lines where trains might
enter.
\item\label{pnt:shared} \textbf{Transfers and shared rolling stock}: passenger
transfers along important railway nodes can lead to a train being delayed when
many of its passengers are coming from a delayed train.
Moreover, some services will share rolling stock between trips, creating more
opportunities for delay propagation.
Crew scheduling conflicts can also worsen delay propagation when extreme delays
occur.
\end{enumerate}

Our approach aims to tackle the latter five points, with extrinsic incidents
being the sole unpredictable and hard-to-estimate factor.
The contributions of the latter four points are what we call \textbf{delay
propagation}, a phenomenon wherein minor perturbations can cause major delays on
entire sectors of the network by exacerbating weaknesses in the circulation plan
and the network's structure.
To address this, we emphasize parallel prediction of all trains, with extensive
history as input: rather than proceeding line-by-line or station-by-station as
most previous work does, our model observes the situation of all trains on the
network at once (as well as its history).

Figure~\ref{fig:propa} contains a concrete example of delay propagation on the
French national railway network on a given day.
The images depict a Gaussian model of the density of recorded delays over a
window of 30 minutes at RPs. This Gaussian density is normalized by the density
of RPs, to avoid dense areas (such as Paris) always registering the highest
values due to the density of traffic and RPs.

The figures are taken at four different times of the day, the first image at
8:30~AM registers a high density of delays in the south,
the second sees it move towards the South-East, then Dijon to the South-East of
Paris on the third image.
The last image at 7:30~PM finally shows a very high density of delays in Metz
in North-Eastern France -- it is remarkable that such a minor railway
node has a higher concentration of delays than the entire Paris area, most
likely due to delay propagation.
\begin{table*}
  \centering
  \caption{Sample of \textbf{BREHAT} data}
  \label{tab:brehat}
  \begin{tabular}{l|l|l|l|l|l}
    \toprule
id & time & RP & type & delay & trainNum\\
    \midrule
462175827 & 2018-01-06 09:42:30 & 681247BV & P & -4 & 6920\\
931562731 & 2018-01-04 20:54:24 & 11320933 & P & 111 & 4453\\
147732417 & 2018-01-08 05:20:55 & 715938WS & O & 16 & 220\\
108326080 & 2018-01-01 22:13:22 & 713339RV & P & 7 & 853221\\
  \bottomrule
\end{tabular}
\end{table*}
The full animation shows an even clearer image of the concentration of delays
moving along the described trajectory.
The delays propagating are not carried by any single train, since no train on
that day travels from Perpignan to Metz, and the delay concentration often pauses
at major railway nodes such as Lyon.
The other concentrations of delays on the map do not follow any clear patterns
of delay propagation in this example, and do not seem to interact with the main
pole of delays shown in the map.

\subsection{Setting}
The French National Rail Network~(NRN) contains about 10,000
\textbf{remarkable points}~(RP), each of which are various types of important
railway nodes: train stations, junctions, bifurcations, train depots, signals,
\ldots{}
All stations are RPs, but not all junctions are RPs, making it a
high-level description of the network which does not account for individual
tracks or switches.
These points can be considered as nodes of a graph of the rail network, with
two remarkable points being connected with an undirected edge if one can be
reached from the other.
The spacing between two consecutive RPs is very heterogeneous, ranging from a
few kilometers within Paris to over 90~km on low-density railways in rural
areas.

About 6,000 of these RPs are equipped to communicate with trains when they
travel through them, creating a database of all train passages at these RPs.
On any given weekday, about 350,000 such events are recorded at \textit{SNCF
Réseau}, France's national railway network manager.
This database provides a rough description of the NRN's state at any given
moment, positioning each train on an edge or node of the network's graph;
however providing no information on the trains' precise position along an edge
nor their speed or state.
This description of the rail network corresponds to the \emph{macro level}
defined by the International Rail Solution's RailTopoModel
standard~\citep{railtopomodel}, which means our approach can be generalized to
any network described with this standard.
\begin{figure}
\includegraphics[width=.95\columnwidth]{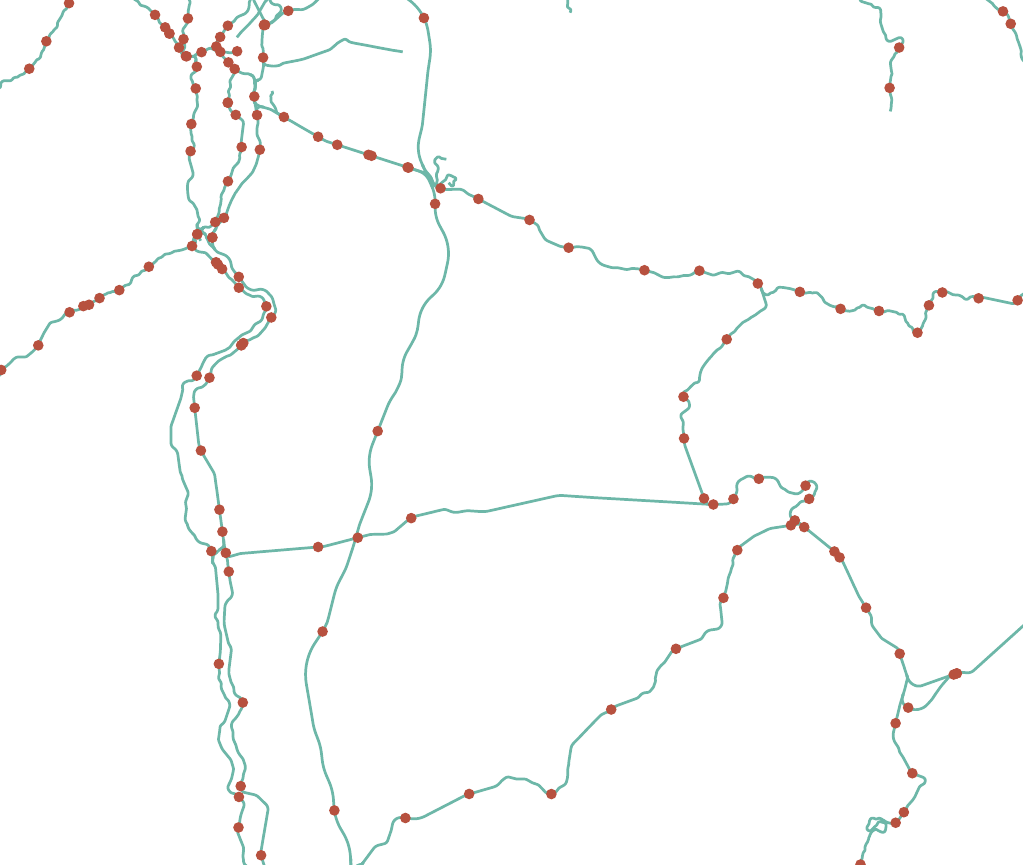}
\caption{Representation of RPs superimposed on railways, area surrounding
Grenoble and Lyon.}
\Description{A map showing train lines with RPs as points along them.}
\end{figure}

On the French NRN, most trains follow a pre-established schedule every day
called the \textbf{circulation plan}, repeated every weekday, with two
alternate versions for Saturdays and Sundays.
Trains are each given a \textbf{train number}, which for most trains designates
a precise itinerary and time schedule, repeated daily.
Some exceptional trains do not follow this scheduled plan and are assigned an
arbitrary number from a given range, and holidays affect the circulation plan
for all trains.

The combination of the NRN state and recently-observed events, as well as
historical data about the circulation plan is enough to make informed
predictions about how trains will behave in the medium term.

\subsection{Data}
\label{sub:data}
Our data consists of raw \textbf{observation events} contained in a
database called \textbf{BREHAT} at \textit{SNCF}. The database is built in
real-time from beacons on the network and the circulation plan, and has been
recorded for the past 10 years.
Each observation in the database contains the following information:
time of observation;
remarkable point where the train flagged the beacon;
observation type, a categorical variable among \textit{O} for
origin, \textit{T} for terminus, \textit{P} for passage (without stopping),
\textit{A} for arrival at a station and \textit{D} for departure from a
station;
train number indicating the train's service;
and measured delay relative to the theoretical circulation plan.

Table~\ref{tab:brehat} shows a sample of \textbf{BREHAT} data.
The measured delay variable can be used to reconstruct the up-to-date
circulation plan by subtracting delays from observation times.
Train numbers contain a lot of information about each train, such as its region
and itinerary, rolling stock type, and whether it is traveling in one direction
or the other along the itinerary.

\begin{figure*}
\centering
\includegraphics[width=\textwidth]{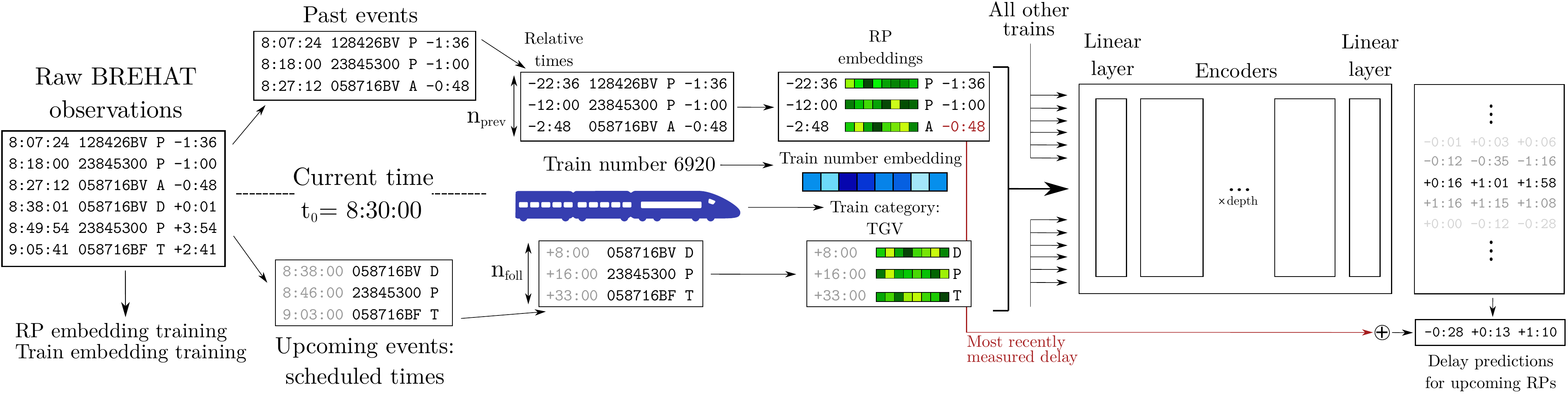}
\caption{Summary of our prediction system}
\label{fig:summ}
\Description{A summary showing the different steps shown in Section 4.}
\end{figure*}

\subsection{Existing approaches}
The \textbf{BREHAT} database is occasionally flawed due to dysfunctional
equipment on the network and inaccuracies in the circulation plan, leading to
both random and repeated drop-out for events.
These inconsistencies in the data, as well as its aforementioned coarseness,
make it impossible to accurately simulate the rail network's evolution. Doing
so would require additional data regarding rail structure at a lower scale than
RPs (signal position and state), physical information about rolling stock
(acceleration and deceleration curves), and extrinsic information such as
driver behavior.
This makes approximation based on past behavior the most rational approach,
preferably modeling all of the delay phenomena listed at the beginning of this
section together rather than separately.

\paragraph{Translation}
Our baseline approach is called \textbf{translation}, which predicts the most
recently observed delay for all following RPs on the train's itinerary.
Translation therefore assumes that a late train will neither catch up its
delay, nor accumulate any further delays, making it a very simple yet
remarkably accurate approach to delay forecasting.
This is the currently used technique at \textit{SNCF} both for passenger
information and traffic control, meaning that phenomena such as race conditions
are never accounted for in predictions as each train is treated individually
and independently of its position or state.
Moreover, since translation only accounts for the \emph{last measured} delay,
a train accumulating a large delay between two RPs will not be detected as late
until it reaches the next RP.

\paragraph{ARMA and ascendancy graphs}
Autoregressive–moving-average (ARMA) models are a widely-used class of models for
time series prediction.
Within SNCF, a first improvement attempt over translation uses control theory and
a conductor/train dynamic~\citep{rousseau2019graphes} wherein a driver attempts
to follow a schedule while the train has an opposing effect, yielding an \(
\text{AR}(2) \) model~\citep{whittle1951hypothesis} for delays:
\begin{equation}
\epsilon(n) = \alpha \epsilon(n-1) + \beta \epsilon(n-2) + \gamma.
\end{equation}
The parameters \( \alpha, \beta, \gamma \) are fit for each train trajectory.
Experiments at the scale of single train lines failed to surpass translation's
general performance, but yielded very positive results for data filtering and
smoothing by using an orthogonal projection and low-pass filtering technique.
This model, which doesn't account for propagation between trains, was compared
to an approach using adjacency graphs and small fully connected and
convolutional neural networks~\citep{rousseau2019graphes}.
An ascendancy graph is built for each train and RP pair \( (t, r) \); and nodes
are created for the \( n \) previous trains going through \( r \). Edges are
weighted according to the time difference between the trains at \( r \), and
the Laplacian matrix for the graph is interpreted as an image by the network.
These experiments failed once more to outperform translation but demonstrated
capability to model interactions between trains.

\paragraph{Bayesian networks}
Another approach previously explored at \textit{SNCF} are Bayesian
networks~\citep{chaker2020modeliser}, which model each pair of train number and
RP \( (t, r) \) as a node in a directed acyclic graph weighted according to the
historical correlation between delay at the target node and at the source node.
The resulting graph, containing over 300,000 nodes, is thinned to only contain
physically meaningful edges: an edge is kept if the target node chronologically
occurs after the source node; and both nodes relate either to the same train,
same RP, or both occur within a given window of time.
The weights of edges designate how the delay from the source train at the
source RP affects the target train at the target RP, and are learned through
historical correlation of these delays.
If we denote \( L(x) \) the delay stored at node \( x \), either the actual
measured delay if it has already occurred, or the prediction if it has not,
then predictions for a given node \( x \) are recursively defined as,
\begin{equation}
L(x) = b(x) + \sum_{y \textrm{ parent of } x} w(y, x) L(y),
\end{equation}
where \( w(y, x) \) are the learned weights through past observed correlation
of delay at \( y \) and \( x \); and \( b \) a learned parameter for each node.
This approach is the simplest and most natural when considering spatio-temporal
propagation of delays, since it should capture delays being propagated from a
train to another by the phenomena described earlier, albeit only in a linear
fashion.

\section{Transformer-based approach}
\label{sec:tran}
Our system performs real-time predictions for all trains on the network, in
parallel, taking into account delay propagation.
For this, we first pre-process \textbf{BREHAT} data to simulate real-time
predictions for training.
This data is also used to create two pre-trained embeddings using auxiliary
neural networks, to embed the RPs and train numbers into meaningful
variables as inputs for the neural network.
The main model, based on the transformer architecture, processes each of these
embedded train vectors as separate tokens of its input and outputs predictions
for all trains at once.
The training set-up for our system is summarized in Figure~\ref{fig:summ}.

\subsection{Preprocessing}
\label{sub:prep}
Given a point in time \( t_0 \), our experimental set-up emulates real-time
conditions from historical \textbf{BREHAT} data, by reconstructing the
theoretical circulation plan used when populating the database.
Events occurring before \( t_0 \) are used as-is, whereas events occurring
after \( t_0 \) are only used to extract information about the theoretical
circulation plan: RPs to be visited, types of observations to be done, and
scheduled time for these events (by adding delay to observation time).
This means there is no information leakage from events occurring after \( t_0
\), even if they were scheduled to occur before \( t_0 \).

Our main model takes as input a list of trains on the network, in the form of
embeddings for each train containing information on their previous itinerary,
and upcoming itinerary.
This information is given to a horizon of \( n_{\textrm{prev}} \) RPs for the
past itinerary and the \( n_{\textrm{foll}} \) RPs for the upcoming itinerary.
For each train, the following information is encoded as a vector and fed to the
main model:
\begin{itemize}
\item The train category: high-speed, regional, freight\ldots\ encoded as
one-hot
\item The embedding of the train number (see Subsection~\ref{sub:temb})
\item The embedding of the \( n_{\textrm{prev}} \) previous RPs visited by the
train (see Subsection~\ref{sub:pemb})
\item The time elapsed since each of these RPs were visited
\item The measured delay at each of these RPs
\item The embedding of the \( n_{\textrm{foll}} \) next RPs to be visited, and
their scheduled time of arrival
\item The observation type of past and upcoming RPs encoded as one-hot
\item Exogenous variables providing context for predictions: day of week, time
of day, and number of trains on network.
\end{itemize}

All trains currently traveling on the network are included as inputs to the
model, but also those that have arrived less than the \( h_{\textrm{arr}} \)
arrival horizon ago, or will leave in less than the \( h_{\textrm{dep}} \)
departure horizon.
Including these trains has several benefits, it allows the model to use data
about trains that have already arrived to better understand the state of the
network.
Including arrived trains helps understand delay propagation through traffic
regulation decisions and shared rolling stock and transfers as described in
points (\ref{pnt:regu}) and (\ref{pnt:shared}) in Section~\ref{sec:prop}.
Including trains that depart soon produces predictions for trains
that haven't left yet, which is important for passenger information and
regulation planning purposes, given these trains may also suffer delays through
propagation.

In the case of trains that have not yet departed or already arrived (or
recently departed or arrived), it is necessary to pad the information on past
or future RPs beyond \( n_{\textrm{prev}} \) or \( n_{\textrm{foll}} \).
For this purpose, we introduce two stand-in RPs labeled \texttt{preDeparture}
and \texttt{postArrival} to fill in embeddings.
All other information (observed delay, time elapsed since RP, \dots) for these
stand-in RPs is arbitrarily set.

\subsection{Remarkable point embedding}
\label{sub:pemb}
Remarkable points given as input to the model as part of the train's itinerary
must contain easily-retrievable information about the graph structure of the
NRN for the model to be able to predict whether two given trains will visit
near-by nodes and potentially interact.  This information can be encoded in an
appropriate graph embedding for the NRN, which must account for heterogeneous
travel times along edges and local graph structure.

To this end, we pre-train our RP embedding on a side task which also uses
the graph structure of the network, in this case \textbf{itinerary length
estimation}.
Given two RPs as their embeddings, a small neural network must estimate the
length in minutes of the shortest path connecting them, as well as the number
of RPs on this path.
The embeddings are considered as variables in the gradient descent algorithm
and are thus trained at the same time as the neural network.

For this, we used historical data to build median travel times between two
given RPs in the NRN, regardless of rolling stock, and use a shortest-path
algorithm to create training data. We also reduce the NRN to its connected
components prior to this, as some parts of the network are fully
grade-separated.
It is important this network be small enough to avoid over-fitting, and train
the embeddings so the information is as easily-retrievable as possible by the
main model.
In our experiments, the network used for training is a simple fully-connected
network of depth 2 and hidden dimension 64~\footnote{A fully-connected network is
a series of linear transformations alternating with non-linear transformations.
Its depth refers to the number of linear transformations in the network, and the
hidden dimension is the dimension to which the first linear transformation maps
the input vector.} using PReLU~\footnote{A ReLU
non-linearity is the function \( x \mapsto \max(x, 0) \). A leaky ReLU is a ReLU
non-linearity with a small coefficient applied to negative values, i.e.\ the
class of functions \( x \mapsto \varepsilon x \ind{x < 0} + x \ind{x \geq 0} \).
A PReLU non-linearity is a parametrized leaky ReLU non-linearity with a learnable
slope parameter \( \varepsilon \), i.e.\ where \( \varepsilon \) is a parameter
that is learned through gradient descent as well.}
non-linearities~\citep{he2015delving}.
Outputs are normalized and training performed using the \( \ell^2 \) loss and the
Adam optimizer~\citep{kingma2014adam}.

\subsection{Train number embedding}
\label{sub:temb}
It is also important to encode the information from train numbers in a
meaningful manner. To this end, we use the side task of \textbf{itinerary
prediction}: given the embedding of a train's number and the embedding of an
RP, a small neural network must predict the next RP visited by the train.
This associates train numbers to RP itineraries, and if this information is
correctly embedded, it allows the main model to look ahead to which trains are
likely to occupy the same RPs in the network.
In order to force generalization and emphasize the importance of train
numbers we add a random drop-out where the current RP is randomly hidden.
Predictions are also sometimes randomly compared to two RPs ahead, or to the
previous RP~\footnote{In particular, this means the output is not a deterministic
function of the inputs.}.

In summary, given a batch of train numbers \( t_i \) and RPs \( r_i \)
respectively visited by those trains and their respective ranks on the trains'
itineraries \( k_i \), define conditionally independent random variables \(
x_i \) of Bernoulli law of probability \( p \); and random independent
variables \( n_i \) taking values \( -1, 1 \) and \( 2 \) with probabilities \(
q_{-1}, q_1 \) and \( q_2 \) respectively.
Denoting the cross-entropy loss \( \mathcal{L} \), the batched loss used for
the model \( f \) is then defined as,
\begin{align}
l\left((t,r), f\right) &= \sum_i \mathcal{L} \left( f(t_i, x_i \cdot r_i),
R(t_i, k_i+n_i) \right) \\
\textrm{where } R(t_i, k_i) &= r_i, \notag
\end{align}
and where \( R \) contains the RPs visited by each train with padding:
\begin{equation}
R(t, k) =
\begin{cases}
\texttt{preDeparture} &\textrm{if } k < 0 \\
\texttt{postArrival} &\textrm{if } k > N_{\max}(t) \\
	k^{\textrm{th}} \textrm{ RP visited by } t & \textrm{otherwise.}
\end{cases}
\end{equation}
The labels \texttt{preDeparture} and \texttt{postArrival} denote two arbitrary
vectors that designate virtual RPs that act as placeholders before a train starts
its itinerary of after they arrive.

In our experiments, we set \( p = 0.15 \) and the vector \( (q_{-1}, q_1, q_2) =
(0.07, 0.75, 0.18 ) \) and use the Adam optimizer~\citep{kingma2014adam}.
The network is made of two fully-connected layers of dimension \( 256 \), with
a PReLU non-linearity (see definition in previous section).

Visualizations and interpretations for our embeddings are presented below in
Subsection~\ref{sec:emb}, and help set hyperparameters for the embeddings
dimension and avoid over- and under-fitting. For instance, setting the
dimension for RP embeddings to \( d_\textrm{RP} = 16 \) gives our embeddings
too much entropy and the \(k\)-closest neighbors for a given RP are spread out
across France: the model has over-fit the embeddings and they no longer contain
geographical information.
Inversely, setting this dimension to \( 4 \) lost too much information and made
the itinerary length prediction model ineffective.

\subsection{Embedding interpretability}\label{sec:emb}
\begin{figure}
\centering
	\subcaptionbox{Train number embeddings for trains along 4 distinct
	itineraries, with trains at different times along the \( x \)-axis and
	embedding dimensions in the \( y
	\)-axis.}{\includegraphics[width=4.2cm]{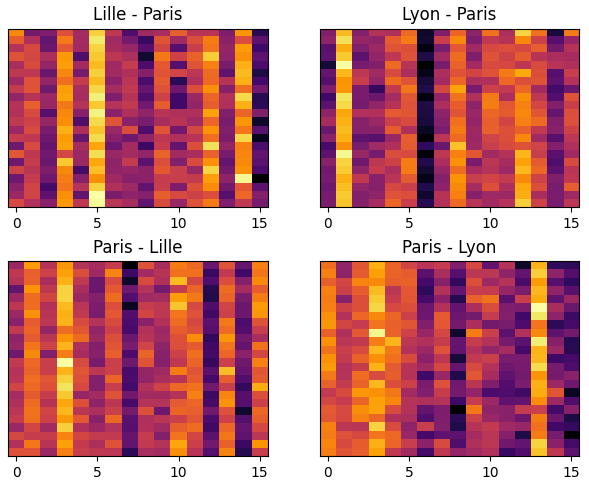}} \hspace{.5cm}
	\subcaptionbox{\( k = 6 \) nearest neighbors (in red) to the Grenoble
	train station remarkable point (in green) in the embedding space.}{\includegraphics[width=3.5cm]{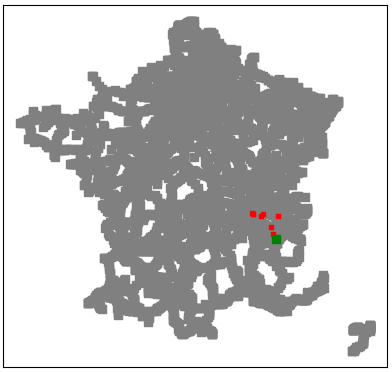}}
	\subcaptionbox{On the left, the distance of remarkable points from Paris
	according to a color scale. On the right, the result of a linear
	regression on the remarkable point embeddings to recover this distance.}{\includegraphics[height=4cm]{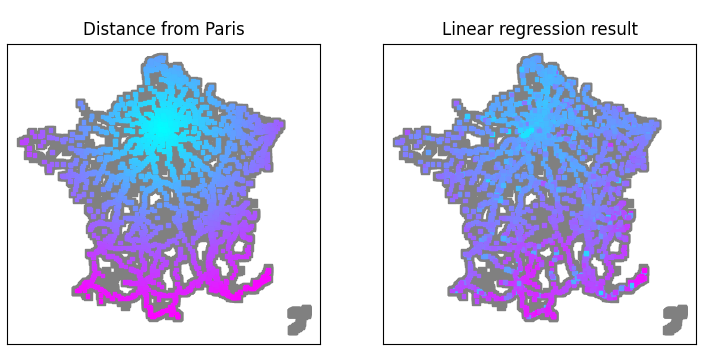}}
\Description{An image showing the embeddings of train numbers as lines with
colored features, similar in each trajectory, the k-nearest neighbors of an RP
are close to it, and a successful linear regression for distance from Paris
from embeddings using color to describe distance}
\caption{Three visualizations of interpretability of our embeddings.}\label{fig:embs}
\end{figure}

In order to avoid adopting a black box end-to-end approach, as well as to aid
with selection of hyperparameters (as explained in the previous subsection), we
verify the interpretability of the trained embeddings on three different
techniques illustrated in Figure~\ref{fig:embs}:
\begin{itemize}
\item Train numbers that travel along the same itinerary at different times
should have similar embeddings, shown here by plotting embeddings for each
train on a specific itinerary at a different time on each line.  Columns are
the different components of the \( d_\textrm{train} = 16 \) dimensions in
the embedding, and lines are different train numbers for the same itinerary.
\item Near-by RPs should have similar embeddings, illustrated here using the
	\(k\)-nearest neighbors (shown in red) for the Grenoble train station RP
	(shown in green) in the embedding space: this ensures that no
	over-fitting has taken place and that near-by RPs for railway distance
	are indeed near-by in the embedding space.
\item Physical location of RPs should be easily extractable from their
embeddings, as illustrated here using linear least-squares regression to
recover geographical distance from Paris from the embeddings.
\end{itemize}

All of these illustrations provide good reasons to believe the embeddings
faithfully encode information about the rail network, and the trains'
schedules.
They were also very useful in setting the embeddings' hyperparameters such as
dimension and training parameters, by providing an \textit{a priori} measure of
their effectiveness without re-training a complete transformer model.

\subsection{Transformer encoder model}
\label{sub:trans}
Our main model is the succession of a linear layer, a transformer encoder
stack, and another linear layer.
The details for the architecture of transformer encoders are explained in detail
in \citet{vaswani2017attention}.
The first linear layer shapes each train's embedded vector into the right
dimension for the encoders.
Each train's vector after this layer is used as a parallel input to the encoder
stack, similarly to a token in a sentence fed to an NLP model.
After the stack of encoders, another linear layer converts the output to the
right dimension for the delay predictions.
Subsection~\ref{sec:attn} provides more insight into the behavior of
self-attention in our task and the rationale behind the use of this
architecture.

The output, of dimension \( n_{\textrm{trains}} \times n_{\textrm{foll}} \),
represents predictions for the next RPs for all trains.
Outputs for the \texttt{postArrival} RP are discarded when computing the
batched loss, since they do not represent real measured delays.
All numeric inputs (times and delays but not embeddings) to the model and all
model outputs are normalized using the empirical mean and standard deviation of
the training data, in order to accelerate training.
Before normalizing, we also convert all delays to their square roots and then
take the square of all output delays before un-normalizing: by working in
square root space, the model is less sensitive to big delays, which avoids few
very late outlier trains taking more importance than the rest.
Therefore, if the delays as input are \( d_{-n_\textrm{prev}}, \ldots, d_{-1}
\) and times are \( t_{-n_\textrm{prev}}, \ldots, t_{n_\textrm{foll} - 1} \),
they will be transformed by \( f \) before and \( f^{-1} \) after the first and
final linear layers of our model:
\begin{align}
	f(x)      & = \textrm{sgn}(x) \sqrt{|x|} \\
	f^{-1}(y) & = \textrm{sgn}(y) y^2 = |y| \; y.
\end{align}
Our experiments also present a model using the logarithm and exponential
functions in place of the square root here, and another using the identity
function for \( f \) and \( f^{-1} \).

Finally, rather than using the output directly as a prediction, we add the
translated delay to all predictions. The translated delay is the last measured
delay for the train, meaning that our model predicts the difference between the
last measured delay and the real delay.
We chose to clip predictions that would lead for an unreached RP to
be predicted in the past:
we apply a ReLU non-linearity~\footnote{A ReLU non-linearity is the function \( x
\mapsto \max(x, 0) \).} to the prediction added to the scheduled
remaining time, before subtracting the scheduled time again; effectively
clipping all predictions to the current time.
This post-processing step is only used when testing the model but not during
training, in order to still correct these harmful predictions and improve
performance -- doing so during training would prevent the model from learning
from these mistakes through gradient descent.

To summarize, given a train \( i \) and the index of an upcoming RP \( j \leq
n_{\textrm{foll}} \), denote \( y_{i,j} \) the model's raw output, \( \mu,
\sigma \) the mean and standard deviation of training data, \( t_{i,-1} \) the
last measured delay for the train, and \( s_{i,j} \) the scheduled remaining
time to the RP (potentially negative when the train is late), then our
prediction is,
\begin{align}
&\sigma f^{-1}(y_{i,j}) + \mu + t_{i,-1} \quad &\textrm{when
training} \\
\max \Big( & \sigma f^{-1}(y_{i,j}) + \mu +
t_{i,-1} + s_{i,j}, 0 \Big) - s_{i,j} \quad &\textrm{when evaluating.}
\end{align}

\subsection{Self-attention between trains}\label{sec:attn}
\begin{figure}
\centering
\includegraphics[height=5.5cm]{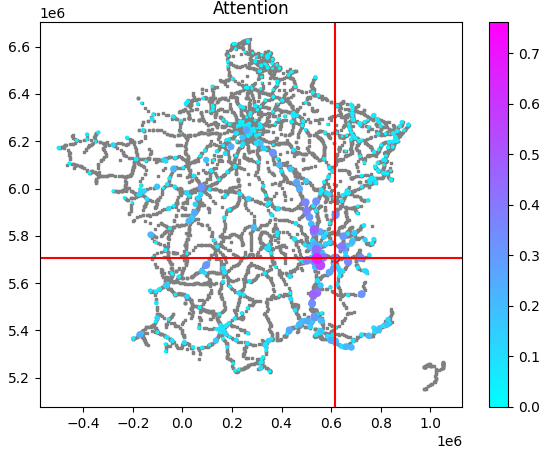}
\Description{Image showing dots for each train sized and colored according to
the self-attention of a particular train.}
\caption{Self-attention in the first encoder for predicting delays for a
Grenoble-Paris TGV shortly after leaving.}\label{fig:attn}
\end{figure}

There are two strong rationales for using a transformer-based architecture for
this task. The first is that we are confronted with a long list of input
tokens, of variable length: the main alternative would have been recurrent
neural networks (such as bi-directional LSTMs), but this would raise questions
about token ordering and memory length to process over 3,000 train tokens as
input.

The second rationale is the attention mechanism, which allows the model to
focus solely on relevant trains when making a prediction for a given train.
Self-attention works by transforming each input token \( x \) into query \( Qx
\), key \( Kx \), and value \( Vx \) vectors in each attention
head~\citep{vaswani2017attention}.
Then, each token \( y \) is weighted relative to \( x \) through the scalar
product \( (Q x) \cdot (K y) \). These weights are normalized using a soft-max
layer and then these scores used to produce a convex combination of each
token's value vector,
\begin{equation} A(x) = \sum_y \mathcal{S} \left( (Q x) \cdot (K y) \right) V y,
\end{equation}
which will constitute the output for token \( x \), where \( \mathcal{S} \) is
the soft-max function relative to the weights.
\begin{table*}
  \caption{Results on the preliminary and main experiments}
  \label{tab:results}
  \begin{tabular}{l|l||l|l|l|l}
    \toprule
& Preliminary & \multicolumn{4}{|c}{Main experiment} \\
& MAE \((\downarrow\)) &
MAE \((\downarrow\)) & MSE \((\downarrow\)) & Incident (\(\uparrow\)) & Service \((\uparrow\))\\
    \midrule
Translation &      2.4 &  5.6 & 22.7 & 77.1 & 76.3 \\
Bayesian network & 2.1 &   -- &   -- &   -- &   -- \\
\midrule
Transf. \( d = 512 \), no-norm &  \textbf{1.7} & 4.2 & 15.2 & 80.1 & 77.0 \\
Transf. \( d = 512 \), log-norm &  -- & 4.1 & 15.1 & 79.8 & \textbf{77.1} \\
Transf. \( d = 1024 \), sqrt-norm &  -- & \textbf{3.7} & \textbf{14.3} & \textbf{81.2} & \textbf{77.1} \\
  \bottomrule
\end{tabular}
\end{table*}

We are interested in the normalized weights \( \left( \mathcal{S}((Q x) \cdot
(K y)) \right)_y \), which constitute a probability distribution over all
trains in the input, modeling the way they affect the train \( x \) in each
self-attention layer.
We therefore chose to monitor alongside our performance metrics the attention
weights for the first encoder in our stack.
This information is extremely valuable since it roughly encodes how delays
propagate: if a model pays particular attention to a set of trains when making
a prediction for a particular train, this most likely means that these
trains are likely to impact the second train.

While this cannot be quantitatively evaluated,
visual analysis of snapshots of the model's attention yields very interesting
results.
Figure~\ref{fig:attn} shows the attention for a high-speed train leaving from
Grenoble to Paris, situated at the intersection of the two red lines.
Each colored point represents another train on the network at the time, whereas
gray points are RPs.
The size and color scale of points quantifies the weight of the self-attention
vector when evaluating the train leaving Grenoble, hence how important they are
in the first encoder block of the model for predicting the delay of the train
from Grenoble to Marseilles.
The train is about to join the high-speed line from Marseilles to Paris, and
the figure shows the model's attention is focused on other trains coming from
Marseilles which are already on the high-speed line.
This is likely what a human operator would also predict, since these trains are
the trains that will also be traveling on the high-speed line between Lyon and
Paris, demonstrating in this example that the model is capable of distinguishing
relevant from irrelevant trains when making a prediction.

These self-attention figures can be of independent interest to train operators in
better understanding the source of delay predictions, and more generally the
mechanisms underlying delay propagation.

\section{Experiments}
\label{sec:res}
We present two experiments, one small preliminary experiment trained on two
months of data to compare our approach to Bayesian networks; and another
large-scale experiment trained on two and a half years of data and tested on
three months of data. This distinction is made due to the conclusive
results of the preliminary experiment, and computation considerations for the
training of Bayesian networks.
All of our models are implemented using the PyTorch library.

We clean \textbf{BREHAT} data for simple inconsistencies such as duplicate
observations or permutations in the order of events.
We remove all local passengers trains for the Île-de-France region (around
Paris), as they constitute a large proportion of data (about a third of
observations, due to traffic and RP density) but are mostly grade-separated
from the rest of traffic.
These trains can thus be treated separately, reducing the amount of tokens as
input to our model.
Our final database of observation events contains about 142M events for the
three years, which averages to 142,000 events a day and 7,500 trains a day.
We then create regularly-spaced training points as described in
Subsection~\ref{sub:prep} for each day between 6 AM and 11 PM, 15 minutes apart
for training days and 4 minutes apart for test days.
Appendix~\ref{app:data} describes data released for reproducibility purposes.

The number of trains as input to our model varies on a normal weekday between
1,000 in the evening and 3,000 during morning and afternoon peak hours.
The average delay for observations is 5 minutes, the median 0 minutes and the
standard deviation 27 minutes: most trains are on time (or early), but late
trains introduce a lot of variation in measured delays.

We set \( n_\textrm{prev} = 10 \) and \( n_\textrm{foll} = 40 \), and
the depth of our encoder stack is \( 2 \), as experiments with deeper stacks
yielded no major improvements.
Our transformer model has dimension \( d_\textrm{model} = 1024 \), a
feed-forward dimension of \( d_\textrm{ff} = 4096 \), \( h = 2 \) heads in each
self-attention layer, and a dropout probability of \( P_\textrm{drop} = 0.1 \)
using notation from~\citet{vaswani2017attention}.
Our chosen embedding dimensions are \( d_\textrm{RP} = 12 \) and \(
d_\textrm{train} = 16 \), and were set according to the interpretability
visualizations explained in Subsection~\ref{sub:temb}.
We use horizons \( h_\text{arr} = 90 \) minutes and \( h_\texttt{dep} = 12 \)
minutes, chosen to minimize the number of trains while conveying enough
information about recently-arrived or departed trains.

Our model is trained using the Adam optimizer~\citep{kingma2014adam} at a
learning rate of \(5\times10^{-5}\) and a batch size of 32, using the L1
loss.
All models are trained on a system using a single NVIDIA Tesla T4 GPU.
The main model was trained over the course of approximately 24~hours.

\subsection{Evaluation}
The two first presented metrics are Mean Average Error (MAE) and Mean Square
Error (MSE).
Mean average error refers to the average \( \ell^1 \) error, i.e. \( \frac{1}{N}
\sum_i \left| \text{predicted}_i - \text{real}_i \right| \) where \( i \) varies
over the \( N \) observations of the experiment.
Mean squared error refers to the average \( \ell^2 \) error, i.e. \( \frac{1}{N}
\sum_i {\left( \text{predicted}_i - \text{real}_i \right)}^2 \).
We use on top of this two non-differentiable \textbf{passenger information
reliability} metrics which are already in use within \textit{SNCF}:
MSE will tend to emphasize errors on large, unpredictable delays, whereas MAE has
a more uniform focus on all types of errors.

We use on top of this two non-differentiable \textbf{passenger information
reliability} metrics which are already in use within \textit{SNCF}:
\renewcommand{\labelenumi}{\roman{enumi}}
\begin{enumerate}
\item the \textbf{incident metric} requires our model to make a
prediction for the rest of the train's visited stations within 10 minutes of
the first measured delay of at least 5 minutes ;
\item the \textbf{service metric} requires our model to make a
prediction for each of the train's visited stations 30 minutes before being
scheduled there, for all trains.
\end{enumerate}
Both of these metrics only make sense for passenger trains, since other trains
do not stop at stations, and the incident metric only applies to trains that
record at least 5 minutes delay at any point on their trip.
The metrics are given as a percentage of predictions accurate within 5 minutes.
In-use systems using human operators and translation score between \( 60 \% \)
to \( 70 \% \) for these metrics.

\subsection{Results}
\label{sub:results}
The results for the preliminary and full experiment are listed in
Table~\ref{tab:results}.
We present two additional, earlier iterations of the transformer model on our
main experiment: both use encoder dimension \( d_\textrm{model} = 512 \), one
uses the identity function and the other the logarithm function when pre and
post-processing delays and times as explained in Subsection~\ref{sub:trans}.

The preliminary experiment delivers a very decisive result concerning the
transformer approach. While Bayesian networks were able to account for some of
the translation's shortcomings, the transformer approach clearly has an edge
over both.
Since translation captures non-propagated delays, any gains over translation are
gains due to delay propagation prediction.
Therefore, while the absolute difference of Bayesian networks and transformers
compared to translation might seem modest, the relevant indicator is the relative
gain: the transformer sees more than double the improvement over translation that
Bayesian network had.

The main experiment shows further gains in the MAE metric for our main
transformer model, with some improvement between the first and final iterations.
We also note a benefit to using the square root space when dealing with delays
and times in our model, especially on the MSE metric.

We were surprised to notice that the service metric is somehow less improved
between iterations of the transformer, even when compared to translation.
Our interpretation of this result is that most trains either run on-time and
will thus automatically be counted as correct guesses (the \( 76.3 \% \) for
translation), or accumulate a great deal of unpredictable primary delay right
before the station (the remaining \( 22.9 \% \)).
A better metric for propagation is the incident metric, where the model is
allowed to measure a first delay on the train before making predictions.
It might seem counterintuitive that this metric is measured higher than the
service metric for all models, but this can be explained by the fixed horizon
of the service metric: many of the predictions for the incident metric will be
done less than 30 minutes before a station.
This metric sees a large improvement between models and over translation,
indicating our model is capable of modeling propagation between trains,
specifically how a late train will increase its delay through the various
effects listed in Section~\ref{sec:prop}.

\section{Conclusion}
We present the phenomenon of delay propagation at the scale of the national
rail network, and detail several factors contributing to it.
We contribute a system that accounts for this propagation when forecasting
delays for trains, based on the transformer architecture and using pre-trained
embeddings.
This technique can be applied to any rail network using the RailTopoModel
standard~\citep{railtopomodel}, which is the case of several European countries'
national networks.
We believe the work could be applied at different scales, such as a
mesoscopic scale, however this would imply a moderate to high computational
overhead.
We also offer a rationale for the use of self-attention and massively parallel
predictions rather than per-train forecasting as suggested in previous work,
along with visual interpretations for the result of self-attention produced by
our model.
This model surpasses by a wide margin both our baseline, translation, and
various other techniques attempted in previous work.

\begin{acks}
The authors would like to thank Vianney Perchet and Francis Bach for their
precious supervision and feedback.
We would also like to thank Amine Chaker and Tom Rousseau for sharing their work
with us.
Finally, the authors thank Bertrand Houzel and Loïc Hamelin at DGEX Solutions
for making this work possible.
\end{acks}

\bibliographystyle{ACM-Reference-Format}
\bibliography{}

\appendix
\section{Reproducibility}
\subsection{Data preparation}
Data cleaning is mentioned in Section~\ref{sec:res}, our preparation consists
in removing duplicate observations that regularly occur in \textbf{BREHAT}
data, and permuting observations in the wrong order.
Duplicates can be detected thanks to an extra rank variable in \textbf{BREHAT}
in \textbf{BREHAT} data: each RP along a train's path is assigned a rank from 1
to the number of expected observations.
These ranks are discontinuous, i.e.\ there are many ranks missing, and are not
unique per observation: when a train arrives and departs from a train station,
there are two distinct observations with the same rank but different
observation types.
We first re-order ranks when they are disordered, by using observation times to
correct mistakes in said order. These errors are most often due to an imperfect
circulation plan being used to create \textbf{BREHAT} data.
We then filter all duplicate observation that have same rank, observation
type, RP and train number for a given day.

\subsection{Data}\label{app:data}
The two following data is published alongside the paper~\citep{data}:
\begin{itemize}
\item Raw \textbf{BREHAT} data for the second week of 2018, containing all
fields described in Subsection~\ref{sub:data}, as a set of JSON files,
\item Training data generated from said data for the same week, using the
process detailed in Subsection~\ref{sub:prep}, as a set of JSON files.
\end{itemize}
The full dataset cannot be released for industrial competition reasons.
\end{document}